\pdfoutput=1
\documentclass[11pt]{article}

\usepackage[]{acl}

\usepackage{times}
\usepackage{latexsym}

\usepackage{graphicx}
\usepackage{multirow}
\usepackage{array}
\usepackage{booktabs}

\usepackage{amsfonts}
\usepackage{bbding}

\usepackage{bm}
\usepackage{color,colortbl}
\usepackage{stfloats}
\usepackage{subfigure}
\usepackage{enumitem}
\usepackage{booktabs}
\usepackage{nccmath}
\usepackage{multirow}
\usepackage{subfigure}

\usepackage{algorithm}
\usepackage{algpseudocode}
\usepackage[T1]{fontenc}
\usepackage[utf8]{inputenc}

\usepackage{microtype}

\title{Guiding Computational Stance Detection with Expanded \\Stance Triangle Framework}

\author{
Zhengyuan Liu\textsuperscript{$\dagger$},
\ Yong Keong Yap\textsuperscript{$\ddag$},
\ Hai Leong Chieu\textsuperscript{$\ddag$},
\ Nancy F. Chen\textsuperscript{$\dagger$}\\
\textsuperscript{$\dagger$}Institute for Infocomm Research (I$^2$R), A*STAR, Singapore\\
\textsuperscript{$\ddag$}DSO National Laboratories, Singapore\\
\texttt{\{liu\_zhengyuan,nfychen\}@i2r.a-star.edu.sg}
\\\texttt{\{yyongkeo,chaileon\}@dso.org.sg}
}

\date{}

\begin{document}
\maketitle

\begin{abstract}
Stance detection determines whether the author of a piece of text is in favor of, against, or neutral towards a specified target, and can be used to gain valuable insights into social media. The ubiquitous indirect referral of targets makes this task challenging, as it requires computational solutions to model semantic features and infer the corresponding implications from a literal statement. Moreover, the limited amount of available training data leads to subpar performance in out-of-domain and cross-target scenarios, as data-driven approaches are prone to rely on superficial and domain-specific features.
In this work, we decompose the stance detection task from a linguistic perspective, and investigate key components and inference paths in this task. The stance triangle is a generic linguistic framework previously proposed to describe the fundamental ways people express their stance. We further expand it by characterizing the relationship between explicit and implicit objects. We then use the framework to extend one single training corpus with additional annotation. Experimental results show that strategically-enriched data can significantly improve the performance on out-of-domain and cross-target evaluation.
\end{abstract}

\section{Introduction}
\label{sec:introduction}
Stance (and its variant stancetaking) is a concept defined as a linguistically articulated form of social action whose meaning is construed within language, interaction, and sociocultural value \cite{biber1988adverbial,agha2003social,du2007stance,kiesling2022stanceAnualReview}. Its subject can be the speaker in a conversation or the author of a social media post, and its object can be in the form of an entity, concept, idea, event, or claim.\footnote{There are various definitions of stance and stancetaking in pragmatics and sociolinguistics field. In this article, we follow the generic ``stance act'' defined by \citet{du2007stance}.}

The stance detection task in natural language processing aims to predict the stance of a piece of text toward specified targets. Stance detection is commonly formulated as a classification problem \cite{kuccuk2020survey}, and is often applied to analyzing online user-generated content such as Twitter and Facebook posts \cite{mohammad-etal-2016-semeval,li2021pstance}. When given the text and one specified target (i.e., stance object), a classifier is used to predict a categorical label (e.g., \textit{Favor}, \textit{Against}, \textit{None}). Along with social networking platforms' growing impact on our lives, stance detection is crucial for various downstream tasks such as fact verification and rumor detection, with wide applications including analyzing user feedback and political opinions \cite{glandt-2021-tweetCovid}. For example, during the pandemic of COVID-19, it was essential to understand the public's opinion on various initiatives and concerns, such as getting booster vaccinations and wearing facial masks. The insight from stance analysis could help public health organizations better estimate the expected efficacy of their mandates, as well as proactively detect pandemic fatigue before it leads to a serious resurgence of the virus.

\begin{table}[t!]
\centering
\small
\resizebox{\linewidth}{!}
{
\begin{tabular}{p{3.5cm}p{3.1cm}}
\toprule
\multicolumn{2}{l}{\textbf{Text}: Service was slow, but the people were friendly.} \\
\textbf{Aspect}: ``Service'' & \textbf{Sentiment}: Negative \\
\textbf{Aspect}: ``people'' & \textbf{Sentiment}: Positive \\
\midrule
\multicolumn{2}{l}{\textbf{Text}: I believe in SCIENCE. I wear a mask for YOUR} \\ PROTECTION. \\
\textbf{Target}: ``wear a mask'' & \textbf{Stance}: Favor \scalebox{0.9}{\CheckmarkBold} \\
\textbf{Target}: ``{\color{blue}Dr. Fauci}'' &  \textbf{Stance}: Favor \scalebox{0.9}{\CheckmarkBold} \\
\textbf{Target}: ``{\color{blue}no mask activity}'' &  \textbf{Stance}: {\color{red}Favor} \scalebox{0.9}{\XSolidBrush} \\
\textbf{Target}: ``CD Disk'' & \textbf{Stance}: {\color{red}Favor} \scalebox{0.9}{\XSolidBrush} \\
\bottomrule
\end{tabular}
}
\caption{\label{table-intro-example}Two examples of aspect-level sentiment analysis and target-aware stance detection. The incorrect label prediction is highlighted in red. The target with implicit mention is highlighted in blue.}
\vspace{-0.3cm}
\end{table}

While state-of-the-art results have been achieved on text classification by adopting data-driven neural approaches, especially utilizing recent large-scale language backbones \cite{devlin-2019-BERT,liu2019roberta}, stance detection remains challenging; there is a substantial gap between human and machine performance.\footnote{Recent empirical evaluation studies show that large language models (LLMs) can provide reasonable results on various NLP tasks including stance detection. However, the performance of adopting zero-shot and few-shot inference with LLMs is still lower than task-specific fine-tuned approaches, and LLMs require great amounts of computational resources \cite{ziems2023can,bang2023multitask}.}
One challenge comes from the ubiquitous indirect referral of targeted stance objects. When interacting socially online, people express their subjective attitude with brevity and variety: they often do not directly mention the final target, but mention its related entities, events, concepts, or claims. As examples shown in Table \ref{table-intro-example}, unlike aspect-based sentiment analysis, where aspect terms are usually explicitly stated in sentences, targets specified for stance labeling can be flexibly assigned. For instance, in a tweet about COVID-19, while ``Dr. Fauci'' is not mentioned, one can infer that the user stands for him from the support of ``wearing a mask'' and ``science''. 
Therefore, target-aware context understanding requires capturing the relationship of explicitly-mentioned objects and various targets, but existing models lack such capability.

Another challenge stems from limited annotated data for stance detection. When training on a corpus constructed with a small number of targets from a single domain, data-driven approaches cannot generalize well on out-of-domain samples and unseen targets \cite{allaway2020VAST,kaushal2021twt}.
Meanwhile, due to low data diversity and the spurious correlation caused by single target labeling, models are prone to over-fit on superficial and biased features (e.g., sentiment-related lexicon). The strong baselines are observed to solely rely on the input text (e.g., tweets) but neglect the specified target \cite{ghosh2019stance,kaushal2021twt}, and fail to make correct predictions when we change the targeted object. As shown in Figure \ref{table-intro-example}, the classifier always produces the same output \textit{Favor}, even when irrelevant targets such as ``CD Disk'' are indicated.

\begin{figure}[t!]
    \begin{center}
    \includegraphics[width=0.80\linewidth]{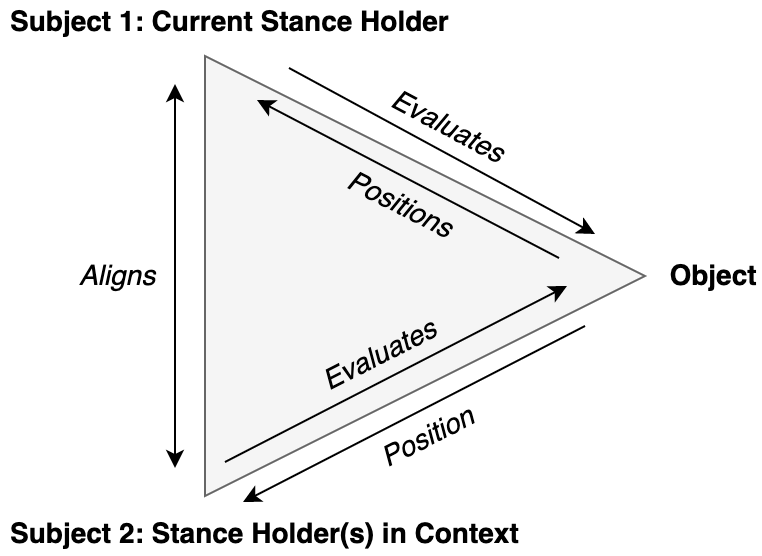}
    \end{center}
    \vspace{-0.1cm}
     \caption{The stance triangle framework proposed by \citet{du2007stance}. Vertices denote the three basic components. Edges denote expression act types.}
    \label{fig-original-triangle}
\vspace{-0.3cm}
\end{figure}

In this work, we investigate solutions for the aforementioned challenges from a linguistic perspective. The pragmatic and linguistics studies provide us with detailed theories of how humans perform stancetaking \cite{du2012Affect,kiesling-etal-2018-interactional}, and help us identify the key components and inference paths for stance analysis. The ``Stance Triangle'' \cite{du2007stance} is one of the most influential and generic linguistic frameworks. As shown in Figure \ref{fig-original-triangle}, it presents three stancetaking acts: a subject (i.e., the stance holder) evaluates an object, positions themselves and others, and aligns with other subjects. While this model covers the important aspects of stancetaking, its broadness leaves the operationalization of stance in practical use cases under-specified \cite{kiesling2022stanceAnualReview}. Regarding stance analysis of social networking platforms, modeling the implication of targets is important, but it is not well-formulated in the triangle framework. Therefore, we expand it by delineating the relationship between explicit and implicit objects, and outline two paths to complete the human-like inference. Aside from using the expanded framework for qualitative analysis, we further utilize it for strategic annotation enrichment, which shows strong potential to improve the robustness and generality of data-driven approaches.
In summary, our contributions of this work are as follows:
\begin{itemize}
\item We make the first attempt to expand the linguistic framework ``stance triangle'' for improving computational stance detection, by characterizing the relationship and labels of explicit and implicit objects.
\item We conduct qualitative analysis following the expanded framework on tweet stance detection, and outline the primary aspects and inference paths.
\item We leverage the proposed framework to enrich the annotation of a single-domain corpus, and empirically demonstrate its effectiveness in improving the performance of out-of-domain and cross-target generalization.
\end{itemize}

\begin{figure*}[t!]
    \begin{center}
    \includegraphics[width=0.83\textwidth]{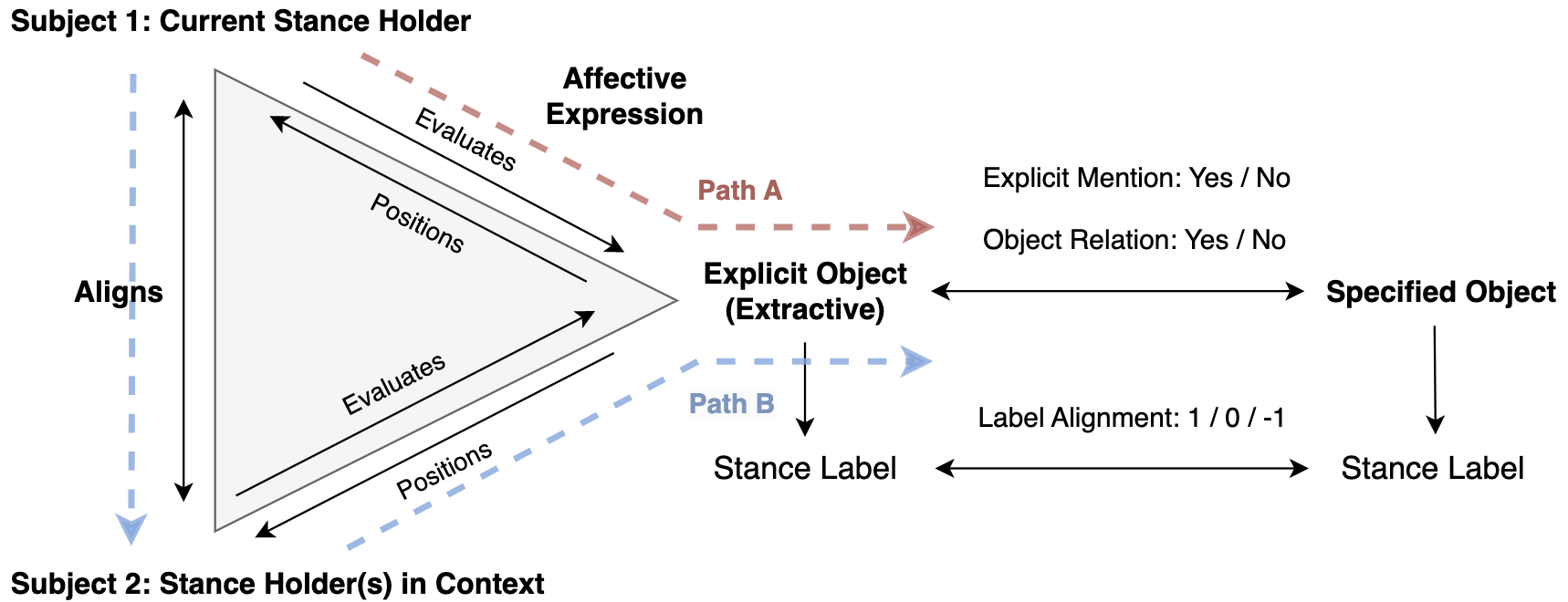}
    \end{center}
     \caption{Overview of our proposed framework expanded on the stance triangle model. The two paths of stancetaking flow are shown by dotted arrow-line in pink and blue color.}
    \label{fig-enhanced-framework}
\vspace{-0.2cm}
\end{figure*}

\section{The Stance Triangle Framework}
\label{sec:expanded_triangle}
In linguistics, stance is originally defined as ``the overt expression of an author's or speaker's attitudes, feelings, judgments, or commitment concerning the message'' \cite{biber1988adverbial}. Sociolinguists further emphasized inter-subjective relations, which refers to how speakers position themselves with their own or other people’s views, attitudes, and claims toward particular objects or ideas in ongoing communicative interaction \cite{haddington2004stance, jaffe2009stance}. In this way, the core concept of stance is embedded with inherently dialogic and inter-subjective characteristics. When other interlocutors in the context indicate a stance on objects, the current speaker may take the same polarity or against their stance in various ways \cite{du2007stance}.

Given that stance is firmly grounded in the communicative interaction and stancetaking is crucial for the social construction of meaning in different discourses, \citet{du2007stance} proposed a holistic analytic framework named ``\textbf{Stance Triangle}'', which describes the communication situation in which two speakers create intersubjectivity through their evaluation of an object. 

As shown in Figure \ref{fig-original-triangle}, in the stance triangle, the three key components regarding stancetaking are located at the vertices, namely the current stance holder (\textbf{Subject 1}), what the stance is about (\textbf{Object}), and the other stance holders in context (\textbf{Subject 2}). The edges are then categorized into three types to reveal expression acts among interlocutors and objects. The evaluation and positioning are both \textbf{Affective Expression} \cite{du2012Affect,kiesling-etal-2018-interactional}.
\textbf{Evaluation} is tied to affect \cite{du2012Affect}. It refers to a person expressing an overt emotional reaction to or displaying an affective orientation toward the object.
\textbf{Positioning} often occurs as a consequence of evaluation. The positioned subject is the one who makes the evaluation toward a specific object. By invoking an evaluation, the subject presents himself/herself as taking a particular affective orientation toward an object via the act of stancetaking.
\textbf{Alignment} is used to highlight the similarities and differences of interlocutors' stance. In communicative interaction, one could find his/her own stance is compared and contrasted with others. As communication goes on, interlocutors may line up affective stances on the shared object. Alternatively, they negotiate their stance and stay differential. Alignment thus becomes an essential part of intersubjectivity in dynamic dialogical interaction.

The stance triangle provides insights into the dialogic nature of stance and serves as an analytical framework to understand how stance is taken in interactions. There are many qualitative studies that are inspired by this framework and applied the fundamental theory to analyze various resources, such as story narration \cite{bohmann2022narration}, and social media text \cite{simaki2018detection}. While previous work demonstrates the success of the stance triangle in linguistics and social constructionism, to the best of our knowledge, few works use the linguistic framework to facilitate computational stance detection.

\begin{table*}[t!]
\centering
\small
\resizebox{0.99\linewidth}{!}
{
\begin{tabular}{p{4.4cm}p{3.5cm}p{3.0cm}p{3.0cm}}
\toprule
\multicolumn{3}{l}{\textbf{Example A}: I believe in SCIENCE. I wear a mask for YOUR PROTECTION.} \\
\multicolumn{2}{l}{\textbf{Explicit Object}: ``wear a mask'' / ``science''} & \textbf{Stance Label}: Favor \\
\midrule
\textbf{Specified Target}: ``{\color{blue}Dr. Fauci}'' & \textbf{Explicit Mention}: No & \textbf{Label Alignment}: 1 & \textbf{Stance Label}: {\color{blue}Favor} \\
\textbf{Specified Target}: ``covid is a hoax'' & \textbf{Explicit Mention}: No & \textbf{Label Alignment}: -1 & \textbf{Stance Label}: Against \\
\textbf{Specified Target}: ``CD Disk'' & \textbf{Explicit Mention}: No & \textbf{Label Alignment}: 0 & \textbf{Stance Label}: None \\
\midrule
\midrule
\multicolumn{3}{l}{\textbf{Example B}: So can unborn children have rights now?} \\
\multicolumn{2}{l}{\textbf{Explicit Object}: ``unborn children''} & \textbf{Stance Label}: Favor \\
\midrule
\textbf{Specified Target}: ``fetus'' & \textbf{Explicit Mention}: No & \textbf{Label Alignment}: 1 & \textbf{Stance Label}: Favor \\
\textbf{Specified Target}: ``{\color{blue}Abortion}'' & \textbf{Explicit Mention}: No & \textbf{Label Alignment}: -1 & \textbf{Stance Label}: {\color{blue}Against} \\
\textbf{Specified Target}: ``Trump'' & \textbf{Explicit Mention}: No & \textbf{Label Alignment}: 0 & \textbf{Stance Label}: None \\
\bottomrule
\end{tabular}
}
% \vspace{-0.1cm}
\caption{\label{table-path-A-example}Two Path-A examples decomposed and annotated based on our expanded stance triangle framework. The original specified target and its label are highlighted in blue.}
% \vspace{-0.1cm}
\end{table*}

\begin{table*}[t!]
\centering
\small
\resizebox{0.99\linewidth}{!}
{
\begin{tabular}{p{4.4cm}p{3.5cm}p{3.0cm}p{3.0cm}}
\toprule
\multicolumn{3}{l}{\textbf{Example A}: Greater is He who is in you than he who is in the world. - 1 John 4:4} \\
\multicolumn{2}{l}{\textbf{Explicit Object}: ``1 John 4:4'' (Quotation)} & \textbf{Stance Label}: Favor \\
\midrule
\textbf{Specified Target}: ``God'' & \textbf{Explicit Mention}: No & \textbf{Label Alignment}: 1 & \textbf{Stance Label}: Favor \\
\textbf{Specified Target}: ``{\color{blue}Atheism}'' & \textbf{Explicit Mention}: No & \textbf{Label Alignment}: -1 & \textbf{Stance Label}: {\color{blue}Against} \\
\midrule
\midrule
\multicolumn{3}{l}{\textbf{Example B}: We remind ourselves that love means to be willing to give until it hurts - Mother Teresa} \\
\multicolumn{2}{l}{\textbf{Explicit Object}: ``Mother Teresa'' (Quotation)} & \textbf{Stance Label}: Favor \\
\midrule
\textbf{Specified Target}: ``the unborn'' & \textbf{Explicit Mention}: No & \textbf{Label Alignment}: 1 & \textbf{Stance Label}: Favor \\
\textbf{Specified Target}: ``{\color{blue}Abortion}'' & \textbf{Explicit Mention}: No & \textbf{Label Alignment}: -1 & \textbf{Stance Label}: {\color{blue}Against} \\
\bottomrule
\end{tabular}
}
\caption{\label{table-path-B-example}Two Path-B examples decomposed and annotated based on our expanded stance triangle framework. The original specified target and its label are highlighted in blue.}
\vspace{-0.2cm}
\end{table*}

\section{Our Proposed Linguistic Framework}
While this triangle model describes the important aspects of stancetaking, its broadness leaves a few limitations when we adopt it to analyze online user-generated content. 
First, the triangle model mainly focuses on dialogic interactions with one shared and explicit object, and does not consider the multiplicity of objects, coreference, and specific quotations on social networking platforms. Meanwhile, the ubiquitous indirect referral of stance targets, and expressions of sarcasm, metaphor, and humor might often require multi-hop reasoning.
Moreover, the stance triangle model only presents the interaction of subject-subject and subject-object, and overlooks the object-object relationship. Therefore, we expand the \textit{object} component to two sub-components: explicit object and specified object, as shown in Figure \ref{fig-enhanced-framework}.

\noindent \textbf{Explicit Object} denotes the explicitly-mentioned object in the text, which the speaker poses a stance on. Its stance label can be obtained from the literal affective expression or alignment between stance holders. Upon this definition, the explicit object can be obtained in an extractive manner from the context, and one piece of text may contain multiple explicit objects, such as the ``wear a mask'' and ``science'' of Example A shown in Table \ref{table-path-A-example}.

\noindent \textbf{Specified Object} denotes the target for predicting the stance label. How the specified object links to an explicit object determines the alignment or dis-alignment of their stance labels.

\noindent \textbf{Explicit Mention} denotes whether the specified target is an explicit object in the context.

\noindent \textbf{Object Relation} indicates whether the stance label of a specified target can be inferred from the explicit object. For instance, regarding the irrelevant target ``CD Disk'', its object relation to the explicit objects ``wear a mask'' and ``science'' is \textit{No}, and that of ``Dr. Fauci'' is \textit{Yes}.

\noindent \textbf{Label Alignment} indicates the relationship between a specified target and the explicit object in stance labeling. 
A value of $1$ means they share the same polarity (e.g., \textit{Favor} vs. \textit{Favor}), and $-1$ means the opposite polarity (e.g., \textit{Favor} vs. \textit{Against}). Moreover, we add the categorical value $0$ when they do not have any stance label correlation (i.e., object relation value is \textit{No}). Therefore, the label alignment can also be used to describe the object relation.

We then outline the two inference paths of the stancetaking flow:
(1) As \textbf{Path-A} shown in Figure \ref{fig-enhanced-framework}, the speaker poses an affective expression on the object from a first-person's perspective.
For Example A shown in Table \ref{table-path-A-example}, the speaker stands for wearing a face mask to protect others in the COVID-19 pandemic. When the specified target is ``covid is a hoax'' which is an implicit object, one can infer that it is related to the explicit object ``wearing a mask'', and their label alignment is $-1$ (i.e., opposite). Thus the stance label of ``covid is a hoax'' is \textit{Against}.

(2) As \textbf{Path-B} shown in Figure \ref{fig-enhanced-framework}, the current stance holder may align or dis-align with other stance holders.
For Example A shown in Table \ref{table-path-B-example}, the speaker quoted a sentence from one chapter ``John 4:4'' of the Bible. This presents the speaker's belief in God, and one can infer that the label of the specified target ``Atheism'' is \textit{Against}.
Moreover, regarding stance analysis of online social networking platforms such as Twitter, the alignment act can be extended to include quotations, re-tweet behavior, and the `Like' button.

The expanded linguistic framework describes the key components, expression acts, and inference flows of stance detection, and it is helpful for qualitative and quantitative analysis, especially for the implicitly-mentioned stance objects.
More importantly, the expanded framework sheds light on the challenging parts of computational stance detection. For instance, some domain knowledge is necessary for reasoning the label alignment between explicit objects and specified targets, and current tweet-related datasets do not provide particular labeling of re-tweet and quotations. This framework paves the way for further research extension.

\begin{table*}[t!]
\centering
\small
\resizebox{1.0\linewidth}{!}
{
\begin{tabular}{p{3.0cm}p{7.3cm}p{1.0cm}<{\centering}p{1.0cm}<{\centering}p{1.0cm}<{\centering}}
\toprule
  \textbf{Corpus} & \textbf{Targets for Stance Labeling} & \textbf{Train}  & \textbf{Valid}  & \textbf{Test} \\
\midrule
  SemEval-16 Task-6 A & Atheism, Climate Change, Feminist Movement, Hillary Clinton, Legalization of Abortion  & 2,914 & - & 1,249 \\
  SemEval-16 Task-6 B & Donald Trump (for zero-shot evaluation) & - & -  & 707 \\
  P-Stance & Donald Trump, Joe Biden, Bernie Sanders & 19,228 & 2,462  & 2,374 \\
  VAST & Various Targets by Human Annotation & 13,477 & 2,062  & 3,006 \\
  Tweet-COVID & Keeping Schools Closed, Dr. Fauci, Stay at Home Orders, Wearing a Face Mask & 4,533 & 800  & 800 \\
\bottomrule
\end{tabular}
}
\caption{\label{table-data-stat}Statistics of the collected stance detection datasets for model training and evaluation.}
\vspace{-0.3cm}
\end{table*}

\section{Theory-inspired Practice: Strategic Annotation Enrichment}
\label{sec:enrich_annotation}
Various corpora with target-aware stance annotation are constructed to facilitate computational solutions for stance detection \cite{mohammad-etal-2016-semeval,allaway2020VAST,li2021pstance,glandt-2021-tweetCovid}.
However, most of them only adopt a simple annotation scheme, where a single target and its corresponding label are provided. Some recent datasets adopt a multi-target annotation \cite{kaushal2021twt}, but the paired target number is limited.

According to our linguistic framework, modeling the implication of a specified object is important. Therefore, we enrich the annotation of one corpus from a single domain by adding multi-target stance labeling on explicit and implicit objects.
We select the data from SemEval2016 Task-6 A ``tweet stance detection'' \cite{mohammad-etal-2016-semeval} as it serves as a benchmark in many previous works \cite{kuccuk2020survey}. As shown in Table \ref{table-data-stat}, it is built on tweets about a set of politics-relevant targets (e.g., politicians, feminism movement, climate change), and each sample only has one specified target with a human-annotated stance label.

We first obtain a sample subset where the specified target is not explicitly mentioned in the text. Next, to obtain explicit objects in an extractive manner, we apply an off-the-shelf constituency parsing model,\footnote{https://demo.allennlp.org/constituency-parsing} and collect all noun phrases in the constituency tree. To reduce extraction noise, we filter out the noun-phrase candidates with some criteria (e.g., being not near the verbs in the sentence, being shorter than 4 characters, and being started with hashtag token and ``$@$user'').

Then linguistic annotators are invited to label the stance polarity on the explicit objects. To reduce superficial target-related patterns and biases from single target labeling, and emphasize object-object relationship, here we propose and adopt an adversarial multi-target strategy, namely selecting the explicit object that shows a stance dis-alignment to the specified target (e.g., ``unborn children'' and ``abortion'' of Example B in Table \ref{table-path-A-example}). This adversarial multi-target labeling can encourage models to condition their prediction more on the given target, as well as learn some correlation between explicit and implicit objects. We obtain 1,500 paired samples, where the original training size is 2,914. Note that we do not introduce any new data to the training set, but enrich the existing corpus. 
Similar to previous work \cite{mohammad-etal-2016-semeval}, our four linguistic annotators participate in the enrichment task (see Appendix Table \ref{table-annotation-detail-appendix} for more details), and the Cohen’s Kappa score calculated for inter-annotator agreement is 0.79 for stance labeling, and according to \citet{uebersax1982generalized}, this score represents a reasonable agreement level.

\section{Experiments on Computational Stance Detection}
\label{sec:experiment_stance_detection}

\subsection{Task Definition}
\label{ssec:task_defenition}
Given $x=\{w_1, w_2,..,w_n\}$ ($n$ denotes the token number) as one input text, and $t=\{t_1, t_2,..,t_m\}$ ($m$ denotes the token number) as the target, the stance detection model is to predict the classification label (e.g., \textit{Favor}, \textit{Against}, \textit{None}). In our experimental setting, we use the 3-class scheme, as the `\textit{None}' label is necessary for practical use cases. Note that in some stance detection corpora, they introduce `\textit{Neutral}' as the third label. To uniform the 3-class labeling for extensive evaluation, we merge `\textit{None}' and `\textit{Neutral}' as one category.

\subsection{Target-Aware Classification}
\label{ssec:classifier_model}
The large-scale pre-trained language models yield state-of-the-art performance in text classification and stance detection \cite{devlin-2019-BERT,kaushal2021twt}. Here we use a Transformer neural network \cite{Ashish-2017-Transformer} as the base architecture, and leverage prior language knowledge by initializing it with the \textit{RoBERTa} \cite{liu2019roberta}.

\noindent \textbf{Target-Aware Encoding}
Since predicting the stance of an input text is dependent on the specified target, previous studies show that conditioning the contextualized representation on the target provides substantial improvements \cite{augenstein-etal-2016-stance,du2017TAN,allaway2020VAST}, thus we concatenate the input text $x$ and the specified target $t$ as one sequence, and use the language backbone to encode it. The input for encoder is formulated as ``<s> $t$ </s> <s> $x$ </s>''.
Then the pooled output of the final layer hidden representation of the first ``<s>'' $v_{\mathrm{enc}}\in\mathbb{R}^E$ (where $E$ is the dimension size) is used as the encoded representation of the target-conditioned input.\footnote{The special tokens vary in different language backbones. For BERT-based models \cite{devlin-2019-BERT}, <s> and </s> are replaced with [CLS] and [SEP], respectively.}

\noindent \textbf{Label Prediction}
To predict the stance label, we feed the encoded representation $v_{\mathrm{enc}}$ to a fully-connected layer and a softmax function to compute the output probabilities:
\begin{equation}
    y^{pred} = \mathrm{softmax}(W^{'}v_{\mathrm{enc}}+b^{'})
\end{equation}
where $W^{'}$ and $b^{'}$ are learnable parameters, and the cross-entropy between gold label $y^{gold}$ and model prediction $y^{pred}$ is minimized as the training loss.

\subsection{Experimental Corpora}
We select several representative stance detection datasets for extensive evaluation, including SemEval-16 Task-6 A and Task-6 B \cite{mohammad-etal-2016-semeval}, P-Stance \cite{li2021pstance}, VAST \cite{allaway2020VAST}, and Tweet-COVID \cite{glandt-2021-tweetCovid}. We use their official train, validation, and test splits. The detailed statistics of these datasets are shown in Table \ref{table-data-stat}. To uniform the label of \textit{None} and \textit{Neutral} from the data perspective, we extend the \textit{None} subset with 20\% size of the training data, by extracting irrelevant objects from random samples, as previous contrastive learning study \cite{gao2021-simCSE}.

In our experiments, models are basically trained on a single-domain corpus (\textit{{\color{black}SemEval-16 Task-6 A}}), and evaluated on multiple test sets. As shown in Table \ref{table-data-stat}, since there are only 5 targets in the single training set of politics-related tweets, testing on different corpora will build the in-domain, out-of-domain, and cross-target evaluation settings \cite{kuccuk2020survey}.
As shown in Table \ref{table-task-type}, \textit{SemEval-16 Task-6 B} contains unseen target ``Donald Trump'', which is used to test the cross-target generalization, and testing on \textit{Tweet-COVID} is both out-of-domain and cross-target.

Moreover, stance label distributions of different targets in our tested benchmark corpora are relatively balanced, and this mitigates the concern of model's over-fitting on ``target-related patterns'' at the evaluation stage.

\begin{table}[t]
\centering
\small
\resizebox{1.0\linewidth}{!}
{
\begin{tabular}{p{2.8cm}p{1.6cm}<{\centering}p{1.8cm}<{\centering}}
\toprule
\textbf{\ Test Set}  & \textbf{In-Domain} & \textbf{Cross-Target} \\
\midrule
\ {\color{black}\textbf{SemEval16 Task-6 A}} & Yes & No \\ 
\ SemEval16 Task-6 B & Yes & Yes \\
\ P-Stance & Yes & Yes \\
\ VAST & No & Yes \\
\ Tweet-COVID & No & Yes \\
\bottomrule
\end{tabular}
}
\caption{\label{table-task-type}Evaluation setting on different test sets. \textbf{\textit{SemEval16 Task-6 A}} is used as the single-domain training corpus for cross-domain and cross-target evluation.}
\vspace{-0.4cm}
\end{table}

\begin{table*}[ht!]
\centering
\small
\resizebox{0.98\linewidth}{!}
{
\begin{tabular}{p{3.1cm}|p{1.4cm}<{\centering}p{1.4cm}<{\centering}p{1.4cm}<{\centering}|p{1.4cm}<{\centering}p{1.4cm}<{\centering}p{1.4cm}<{\centering}}
\toprule
 \textbf{Model}: RoBERTa-base &  \multicolumn{3}{c}{\textbf{In-Domain \& In-Target} (UB.)} & \multicolumn{3}{c}{\textbf{Single Corpus Training}} \\
\textbf{Test Set}  & \textbf{F1} & \textbf{Precision} & \textbf{Recall} & \textbf{F1} & \textbf{Precision} & \textbf{Recall} \\
\midrule
SemEval-16 Task-6 A 	&	0.6849	&	0.6755	&	0.7169	&	0.6849	&	0.6755	&	0.7169	 \\
SemEval-16 Task-6 B 	&	 - 	&	 - 	&	 - 	&	0.4134	&	0.5132	&	0.4389	 \\
P-Stance 	&	0.6344	&	0.6435	&	0.6288	&	0.3454	&	0.4840	&	0.3980	 \\
VAST 	&	0.7375	&	0.7499	&	0.7373	&	0.4079	&	0.4215	&	0.4140	 \\
Tweet-COVID 	&	0.7474	&	0.7534	&	0.7483	&	0.3579	&	0.4334	&	0.4032	 \\
\midrule
\midrule
 \textbf{Model}: RoBERTa-base &  \multicolumn{3}{c}{\textbf{Only Enriched Train Set}} & \multicolumn{3}{c}{\textbf{Adding Enriched Train Set}} \\
\textbf{Test Set}  & \textbf{F1} & \textbf{Precision} & \textbf{Recall} & \textbf{F1} & \textbf{Precision} & \textbf{Recall} \\
\midrule
SemEval-16 Task-6 A & 0.6862 & 0.6774 & 0.7095 & 0.7047 & 0.6912 & 0.7355 \\
SemEval-16 Task-6 B & 0.6439 & 0.6493 & 0.6409 & 0.6885 & 0.6994 & 0.7010 \\
P-Stance & 0.4782 & 0.5175 & 0.4872 & 0.5003 & 0.5152 & 0.5007 \\
VAST & 0.6278 & 0.6488 & 0.6426 & 0.6346 & 0.6783 & 0.6462 \\
Tweet-COVID & 0.5202 & 0.5624 & 0.5349 & 0.5599 & 0.5821 & 0.5752 \\
\bottomrule
\end{tabular}
}\caption{\label{table-result-3-class}Results of the 3-class stance classification on multiple corpora. Macro-averaged F1, Precision, and Recall scores are reported. \textit{UB.} denotes the upper bound result from in-domain and in-target training on each corpus. Results of 2-class macro-averaged scores are shown in Appendix Table \ref{table-result-2-class}. Some examples of model prediction are shown in Appendix Table \ref{table-appendix-examples}.}
\vspace{-0.3cm}
\end{table*}

\subsection{Training Configuration}
Models are implemented with Pytorch and Hugging Face Transformers\footnote{https://github.com/huggingface/transformers}. For fine-tuning on stance detection task, we train the language backbone \textit{RoBERTa-base} \cite{liu2019roberta} with the AdamW optimizer \citep{diederik-kingma-2015-Adam} and batch size $32$. Initial learning rates are all set at $2e^{-5}$, and a linear scheduler ($0.9$ decay ratio) is added. Test results are reported with the best validation scores.

As previous work \cite{mohammad-etal-2016-semeval, allaway2020VAST}, we adopt the macro-averaged F1, Precision, and Recall scores as evaluation metrics. The macro-averaged scheme weighs each of the classes equally and is not influenced by the imbalanced sample number of each class.

\subsection{Experimental Results}
Since we train the model on a single corpus (2.9k samples), testing it on multiple out-of-domain data and various unseen targets poses a challenging task. As shown in Table \ref{table-result-3-class}, compared with in-domain and in-target training on each corpus (which serves as the upper bound for external testing), scores of single-corpus training become much lower, and F1, precision, and recall are all affected significantly. This indicates that the original data only enable the model to achieve reasonable results on in-domain samples and existing targets. In contrast, training with the strategically-enriched annotation (1,500 paired samples) improves the performance substantially and consistently: on the four external test sets, the \textit{RoBERTa-base} model has achieved at least 38\% relative improvement of 3-class labeling. This demonstrates that the model learns more general and domain-invariant features which are useful across different stance detection corpora. Moreover, merging the original data and the enriched set brings further improvement, where at least 45\% relative improvement of 3-class labeling is observed.

For extensive comparison with previous work \cite{mohammad-etal-2016-semeval,li2021targetAug}, aside from the 3-class calculation, we report 2-class macro-averaged scores (i.e., \textit{Favor}, \textit{Against}), where the \textit{None} label is used during training, but discarded in evaluation. As shown in Table \ref{table-result-2-class}, training with enriched data also provides a significant improvement (at least 48\% relative gain), and state-of-the-art cross-target results.

\subsection{Analysis on Target Dependency}
Previous work found that strong baselines often solely rely on the input text but neglect the specified targets \cite{ghosh2019stance,kaushal2021twt}, resulting in poor performance on diverse and unseen targets. Since our model (trained with enriched set) is expected to show better target dependency, we envision that on label-balanced test sets, the distributions of predictions with or without specified targets shall be pretty distinct.

\begin{figure}[t!]
    \begin{center}
    \includegraphics[width=0.99\linewidth]{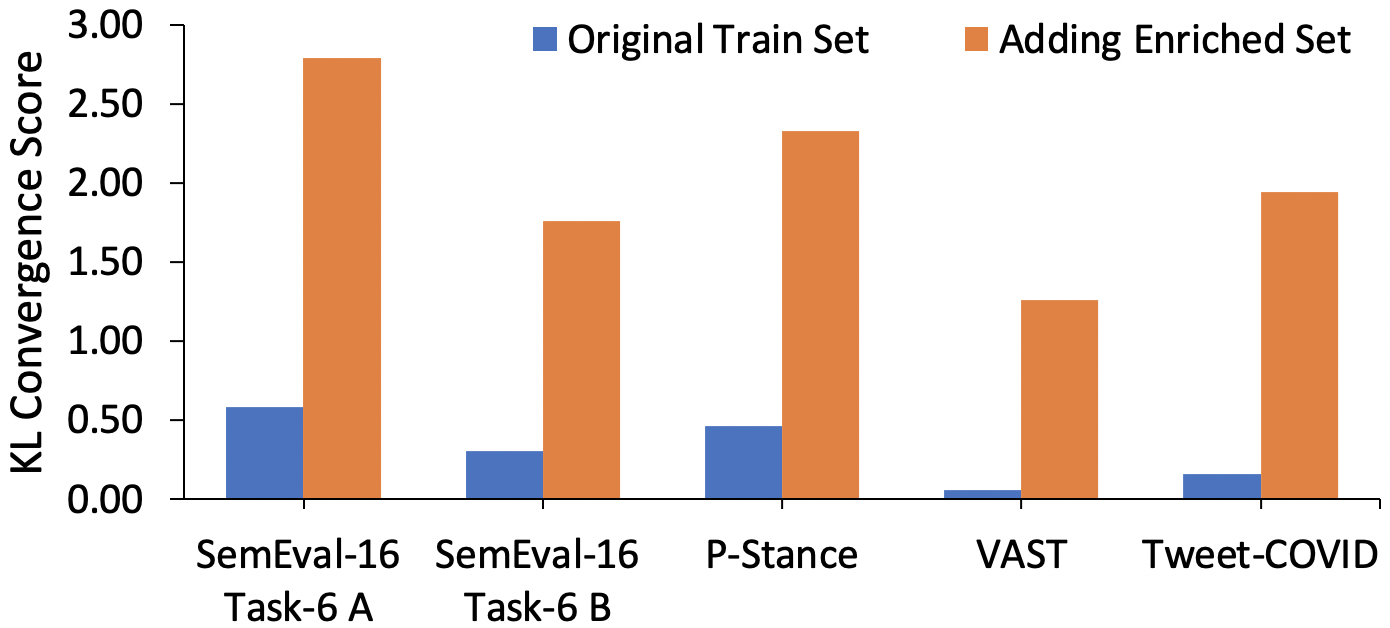}
    \end{center}
    \vspace{-0.2cm}
     \caption{KL divergence comparison of predictions with targets and without targets.}
    \label{fig-KL-dist}
\vspace{-0.2cm}
\end{figure}

Here we conduct a quantitative analysis based on KL divergence as in Equation \ref{EQ-KL}. Given $Q(x)$ is the prediction solely on the input text, and $P(x)$ is conditioned on the specified target, we calculate their KL divergence on the whole test set $\mathcal{X}$ to measure their similarity.
\vspace{-0.2cm}
\begin{equation}
\small
\mathrm{KL}(P||Q)=\sum_{x\in\mathcal{X}}P(x)\log(\frac{P(x)}{Q(x)})
\vspace{-0.1cm}
\label{EQ-KL}
\end{equation}
As shown in Figure \ref{fig-KL-dist}, compared with training on the original set, adding the enriched data results in larger KL divergence values. This empirically shows that model's prediction depends more on the specified targets than the base model.

\subsection{Analysis on Enriched Annotation Size}
\begin{figure}[t!]
    \begin{center}
    \includegraphics[width=0.87\linewidth]{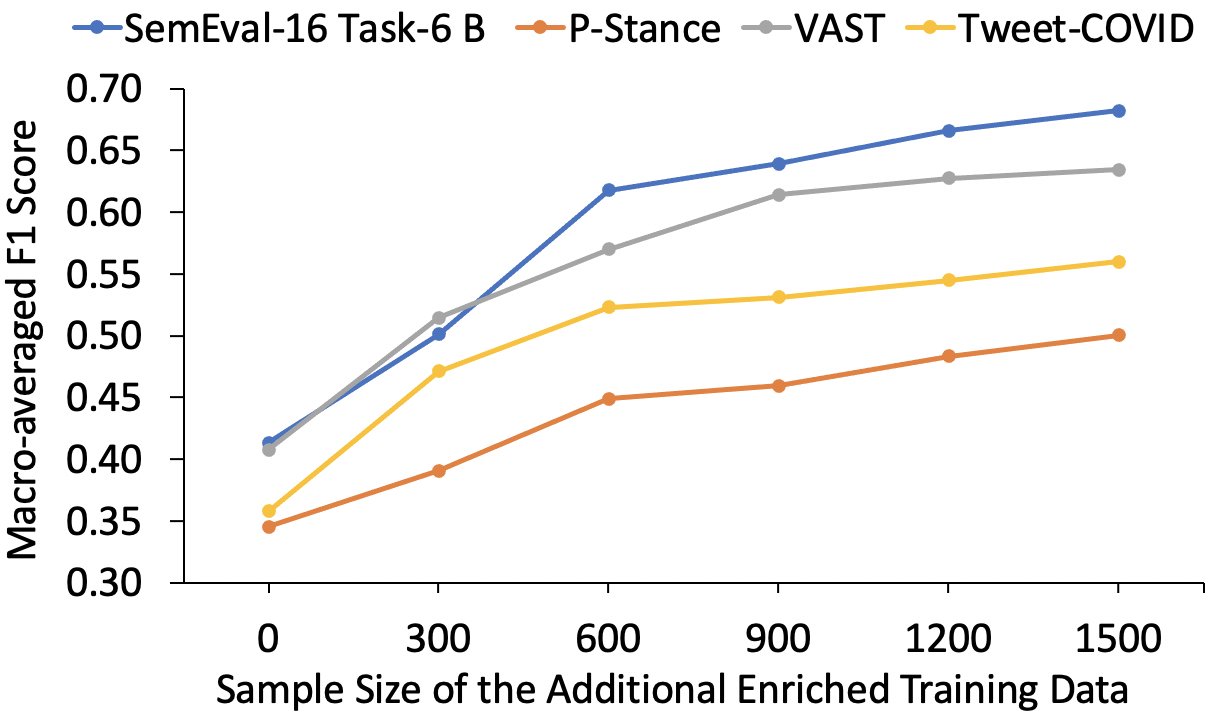}
    \end{center}
    \vspace{-0.2cm}
     \caption{Results on different enriched annotation sizes. Y axis denotes the macro-averaged F1 Score calculated on 3-class prediction, and X axis denotes the additional sample size.}
    \label{fig-enriched-size}
\vspace{-0.2cm}
\end{figure}

We then take an assessment on different sizes of the enriched annotation. As shown in Figure \ref{fig-enriched-size}, compared with training on the original sample set, the performance on out-of-domain and cross-target evaluation can be boosted with 600 enriched samples. We speculate that by leveraging the prior knowledge of pre-trained language backbones, models can learn the general features effectively and efficiently from the enriched data. This demonstrates one advantage of following a linguistic framework for strategic annotation, where models can obtain substantial gain upon limited annotation cost.

\begin{table*}[t!]
\centering
\small
\resizebox{0.97\linewidth}{!}
{
\begin{tabular}{p{3.0cm}|p{1.5cm}<{\centering}p{1.5cm}<{\centering}p{1.5cm}<{\centering}|p{1.5cm}<{\centering}p{1.5cm}<{\centering}p{1.5cm}<{\centering}}
\toprule
 &  \multicolumn{3}{c}{\textbf{Single Corpus Training}} & \multicolumn{3}{c}{\textbf{Adding Enriched Train Set}} \\
\textbf{Test Set}  & \textbf{ATAE} & \textbf{PoE} & \textbf{BERTweet} & \textbf{ATAE} & \textbf{PoE} & \textbf{BERTweet} \\
\midrule
SemEval16 Task-6 A & 0.5604 & 0.6095 & 0.6833 & 0.5676 & 0.6507 & 0.7110  \\
SemEval16 Task-6 B & 0.2744 & 0.4853 & 0.5488 & 0.3297 & 0.6623 & 0.6533  \\
P-Stance & 0.3263 & 0.3720 & 0.4255 & 0.3401 & 0.4939 & 0.4763  \\
VAST & 0.3227 & 0.3656 & 0.3830 & 0.3703 & 0.6220 & 0.5757  \\
Tweet-COVID & 0.3146 & 0.3807 & 0.4707 & 0.4459 & 0.5048 & 0.5658  \\
\bottomrule
\end{tabular}
}
\caption{\label{table-result-3-class-other-models}Various model performance of the 3-class stance classification on multiple corpora. Macro-averaged F1 scores are reported. Results of the 2-class macro-averaged scores are shown in Appendix Table \ref{table-result-2-class-other-models}.}
\vspace{-0.2cm}
\end{table*}

\subsection{Effectiveness across Model Architectures}
We further conduct experiments with other strong baselines in different model architectures and designs:
(1) \textbf{ATAE} \cite{wang2016ATAE}: an LSTM-based model that extracts target-specific features via an attention mechanism.
(2) \textbf{BERTweet} \cite{nguyen2020bertweet}: a BERT-based language backbone that is specially pre-trained on tweet data.
(3) \textbf{Product-of-Expert} (PoE) \cite{clark-etal-2019-PoE}: a de-biasing method that reweights the learned probabilities upon a bias-only model. We train and evaluate these models following the same experimental setting described in Section \ref{sec:experiment_stance_detection}, and their full implementation details are shown in in Appendix Table \ref{table-paramater-detail-appendix}.
As shown in Table \ref{table-result-3-class-other-models} and Table \ref{table-result-2-class-other-models}, the results in most aspects is improved substantially after adding the enriched data, which shows that the strategic augmentation is effective on various architectures. In addition, BERT-based models (e.g., \textit{BERTweet}, \textit{PoE}) show a larger performance gain than the \textit{ATAE}, as they leverage prior language knowledge from pre-training. The comparable result of \textit{RoBERTa-base}, \textit{BERTweet}, and \textit{PoE} shows various backbones can learn domain-invariant features from the enriched data, demonstrating the effectiveness of our adversarial multi-target annotation strategy (Section \ref{sec:enrich_annotation}).

\section{Related Work}
\label{sec:related_work}
\noindent \textbf{Linguistic Studies of Stancetaking}
Stance analysis plays an essential role in measuring public opinion on mass media, conversations, and online networking platforms, particularly related to social, religious, and political issues \cite{agha2003social,kiesling2022stanceAnualReview}. There are many qualitative studies conducted on various resources such as news interviews \cite{haddington2004stance}, twitter and Facebook posts \cite{simaki2018detection}, narrative articles \cite{bohmann2022narration}, and online forums \cite{kiesling-etal-2018-interactional}. Recent works also perform in-depth content analyses on social media images to understand how politicians use images to express ideological rhetoric \cite{xi2020understanding}.
To conduct analyses of stancetaking in a well-formulated manner, many linguistic studies explore the explanation for the semantics, pragmatics, syntactic distribution of lexical items, discourse markers, and syntactic construction across languages \cite{biber1988adverbial,haddington2004stance,lempert2008poetics,jaffe2009stance,du2012Affect}.
From the perspective of social pragmatics, stancetaking on social networks is regarded as a dynamic and dialogic activity where participants are actively engaged in virtual interactions \cite{du2007stance}. Tweets are not viewed as stagnant posts, instead, they become back-and-forth interactions enacted by retweets and comments \cite{chiluwa2015stance,evans2016stance}.

\noindent \textbf{Computational Stance Detection}
Computational stance detection is commonly formulated as a target-specified classification problem. Datasets for stance detection are usually collected from online networking platforms where large-scale user-generated content is available \cite{mohammad-etal-2016-semeval,li2021pstance,allaway2020VAST,glandt-2021-tweetCovid}. Support vector machines (SVM) with manually-engineered features served as the earlier strong baseline \cite{mohammad-etal-2016-semeval}. Then various deep learning techniques such as recurrent neural networks (RNNs) \cite{zarrella2016mitre}, convolutional neural networks (CNNs) \cite{vijayaraghavan2016deepstance}, and attention mechanism \cite{augenstein-etal-2016-stance, du2017TAN,zhou2017connecting} are applied for better feature extraction.
Recently, the Transformer networks \cite{Ashish-2017-Transformer}, especially language backbones \cite{devlin-2019-BERT,liu2019roberta}, boosted the performance on stance detection benchmarks \cite{ghosh2019stance,li2021targetAug}. However, it has been observed that current strong baselines relied heavily on superficial features in existing data, and showed poor performance on unseen-target and out-of-domain evaluation \cite{kaushal2021twt}, and recent work proposed de-biasing methods \cite{clark-etal-2019-PoE} and introduced multi-task learning \cite{yuan2022ssr}.

\section{Conclusions}
In this work, we revisited the challenges of computational stance detection from a linguistic perspective. We expanded a generic linguistic framework the ``stance triangle'', with the relationship and labels of explicit and implicit objects, and characterized various fashions of how humans express their stances. We then followed the framework to strategically enrich the annotation of one benchmarked single-domain corpus. Experimental results showed that the enriched data can significantly improve the performance on out-of-domain and cross-target evaluation. Moreover, our framework paves the way for future research such as assessing explainability of data-driven models.

\section*{Limitations}
All samples used in this work are in English, thus to apply the model to other languages, it will require training data on the specified language or using multilingual language backbones.
Moreover, we are aware that it remains an open problem to mitigate biases in human stancetaking. Of course, current models and laboratory experiments are always limited in this or similar ways. We do not foresee any unethical uses of our proposed methods or its underlying tools, but hope that it will contribute to reducing incorrect system outputs.

\section*{Ethics and Impact Statement}
We acknowledge that all of the co-authors of this work are aware of the provided ACL Code of Ethics and honor the code of conduct. All data used in this work are collected from existing published NLP studies. Following previous work, the annotated corpora are only for academic research purposes and should not be used outside of academic research contexts. Our proposed framework and methodology in general do not create a direct societal consequence and are intended to be used to prevent data-driven models from over-fitting domain-dependent and potentially-biased features.

\section*{Acknowledgments}
This research was supported by funding from the Institute for Infocomm Research (I2R), A*STAR, Singapore, and DSO National Laboratories, Singapore. We thank Yizheng Huang and Stella Yin for preliminary data-related discussions, and Siti Umairah Md Salleh, Siti Maryam Binte Ahmad Subaidi, Nabilah Binte Md Johan, and Jia Yi Chan for linguistic resource construction. We also thank the anonymous reviewers for their precious feedback to help improve and extend this piece of work.

\bibliography{custom}

\begin{thebibliography}{39}
\expandafter\ifx\csname natexlab\endcsname\relax\def\natexlab#1{#1}\fi

\bibitem[{Agha(2003)}]{agha2003social}
Asif Agha. 2003.
\newblock The social life of cultural value.
\newblock \emph{Language \& communication}, 23(3-4):231--273.

\bibitem[{Allaway and Mckeown(2020)}]{allaway2020VAST}
Emily Allaway and Kathleen Mckeown. 2020.
\newblock Zero-shot stance detection: A dataset and model using generalized
  topic representations.
\newblock In \emph{Proceedings of the 2020 Conference on Empirical Methods in
  Natural Language Processing (EMNLP)}, pages 8913--8931.

\bibitem[{Augenstein et~al.(2016)Augenstein, Rockt{\"a}schel, Vlachos, and
  Bontcheva}]{augenstein-etal-2016-stance}
Isabelle Augenstein, Tim Rockt{\"a}schel, Andreas Vlachos, and Kalina
  Bontcheva. 2016.
\newblock \href {https://doi.org/10.18653/v1/D16-1084} {Stance detection with
  bidirectional conditional encoding}.
\newblock In \emph{Proceedings of the 2016 Conference on Empirical Methods in
  Natural Language Processing}, pages 876--885, Austin, Texas. Association for
  Computational Linguistics.

\bibitem[{Bang et~al.(2023)Bang, Cahyawijaya, Lee, Dai, Su, Wilie, Lovenia, Ji,
  Yu, Chung et~al.}]{bang2023multitask}
Yejin Bang, Samuel Cahyawijaya, Nayeon Lee, Wenliang Dai, Dan Su, Bryan Wilie,
  Holy Lovenia, Ziwei Ji, Tiezheng Yu, Willy Chung, et~al. 2023.
\newblock A multitask, multilingual, multimodal evaluation of chatgpt on
  reasoning, hallucination, and interactivity.
\newblock \emph{arXiv preprint arXiv:2302.04023}.

\bibitem[{Biber and Finegan(1988)}]{biber1988adverbial}
Douglas Biber and Edward Finegan. 1988.
\newblock Adverbial stance types in english.
\newblock \emph{Discourse processes}, 11(1):1--34.

\bibitem[{Bohmann and Ahlers(2022)}]{bohmann2022narration}
Axel Bohmann and Wiebke Ahlers. 2022.
\newblock Stance in narration: Finding structure in complex sociolinguistic
  variation.
\newblock \emph{Journal of Sociolinguistics}, 26(1):65--83.

\bibitem[{Chiluwa and Ifukor(2015)}]{chiluwa2015stance}
Innocent Chiluwa and Presley Ifukor. 2015.
\newblock ‘war against our children’: Stance and evaluation in\#
  bringbackourgirls campaign discourse on twitter and facebook.
\newblock \emph{Discourse \& Society}, 26(3):267--296.

\bibitem[{Clark et~al.(2019)Clark, Yatskar, and
  Zettlemoyer}]{clark-etal-2019-PoE}
Christopher Clark, Mark Yatskar, and Luke Zettlemoyer. 2019.
\newblock \href {https://doi.org/10.18653/v1/D19-1418} {Don{'}t take the easy
  way out: Ensemble based methods for avoiding known dataset biases}.
\newblock In \emph{Proceedings of the 2019 Conference on Empirical Methods in
  Natural Language Processing and the 9th International Joint Conference on
  Natural Language Processing (EMNLP-IJCNLP)}, pages 4069--4082, Hong Kong,
  China. Association for Computational Linguistics.

\bibitem[{Devlin et~al.(2019)Devlin, Chang, Lee, and
  Toutanova}]{devlin-2019-BERT}
Jacob Devlin, Ming-Wei Chang, Kenton Lee, and Kristina Toutanova. 2019.
\newblock {BERT}: Pre-training of deep bidirectional transformers for language
  understanding.
\newblock In \emph{Proceedings of the 2019 Conference of the North {A}merican
  Chapter of the Association for Computational Linguistics: Human Language
  Technologies,}, pages 4171--4186.

\bibitem[{Du et~al.(2017)Du, Xu, He, and Gui}]{du2017TAN}
Jiachen Du, Ruifeng Xu, Yulan He, and Lin Gui. 2017.
\newblock Stance classification with target-specific neural attention networks.
\newblock International Joint Conferences on Artificial Intelligence.

\bibitem[{Du~Bois(2007)}]{du2007stance}
John~W Du~Bois. 2007.
\newblock The stance triangle.
\newblock \emph{Stancetaking in discourse: Subjectivity, evaluation,
  interaction}, 164(3):139--182.

\bibitem[{Du~Bois and K{\"a}rkk{\"a}inen(2012)}]{du2012Affect}
John~W Du~Bois and Elise K{\"a}rkk{\"a}inen. 2012.
\newblock Taking a stance on emotion: Affect, sequence, and intersubjectivity
  in dialogic interaction.
\newblock \emph{Text \& Talk}, 32(4):433--451.

\bibitem[{Evans(2016)}]{evans2016stance}
Ash Evans. 2016.
\newblock Stance and identity in twitter hashtags.
\newblock \emph{Language@ internet}, 13(1).

\bibitem[{Gao et~al.(2021)Gao, Yao, and Chen}]{gao2021-simCSE}
Tianyu Gao, Xingcheng Yao, and Danqi Chen. 2021.
\newblock Simcse: Simple contrastive learning of sentence embeddings.
\newblock In \emph{Proceedings of the 2021 Conference on Empirical Methods in
  Natural Language Processing}, pages 6894--6910.

\bibitem[{Ghosh et~al.(2019)Ghosh, Singhania, Singh, Rudra, and
  Ghosh}]{ghosh2019stance}
Shalmoli Ghosh, Prajwal Singhania, Siddharth Singh, Koustav Rudra, and
  Saptarshi Ghosh. 2019.
\newblock Stance detection in web and social media: a comparative study.
\newblock In \emph{International Conference of the Cross-Language Evaluation
  Forum for European Languages}, pages 75--87. Springer.

\bibitem[{Glandt et~al.(2021)Glandt, Khanal, Li, Caragea, and
  Caragea}]{glandt-2021-tweetCovid}
Kyle Glandt, Sarthak Khanal, Yingjie Li, Doina Caragea, and Cornelia Caragea.
  2021.
\newblock \href {https://doi.org/10.18653/v1/2021.acl-long.127} {Stance
  detection in {COVID}-19 tweets}.
\newblock In \emph{Proceedings of the 59th Annual Meeting of the Association
  for Computational Linguistics and the 11th International Joint Conference on
  Natural Language Processing (Volume 1: Long Papers)}, pages 1596--1611,
  Online. Association for Computational Linguistics.

\bibitem[{Haddington et~al.(2004)}]{haddington2004stance}
Pentti Haddington et~al. 2004.
\newblock Stance taking in news interviews.
\newblock \emph{SKY Journal of Linguistics}, 17:101--142.

\bibitem[{Jaffe et~al.(2009)}]{jaffe2009stance}
Alexandra Jaffe et~al. 2009.
\newblock \emph{Stance: sociolinguistic perspectives}.
\newblock Oup Usa.

\bibitem[{Kaushal et~al.(2021)Kaushal, Saha, and Ganguly}]{kaushal2021twt}
Ayush Kaushal, Avirup Saha, and Niloy Ganguly. 2021.
\newblock twt--wt: A dataset to assert the role of target entities for
  detecting stance of tweets.
\newblock In \emph{Proceedings of the 2021 Conference of the North American
  Chapter of the Association for Computational Linguistics: Human Language
  Technologies}, pages 3879--3889.

\bibitem[{Kiesling(2022)}]{kiesling2022stanceAnualReview}
Scott~F Kiesling. 2022.
\newblock Stance and stancetaking.
\newblock \emph{Annual Review of Linguistics}, 8:409--426.

\bibitem[{Kiesling et~al.(2018)Kiesling, Pavalanathan, Fitzpatrick, Han, and
  Eisenstein}]{kiesling-etal-2018-interactional}
Scott~F. Kiesling, Umashanthi Pavalanathan, Jim Fitzpatrick, Xiaochuang Han,
  and Jacob Eisenstein. 2018.
\newblock \href {https://doi.org/10.1162/coli_a_00334} {Interactional
  stancetaking in online forums}.
\newblock \emph{Computational Linguistics}, 44(4):683--718.

\bibitem[{Kingma and Ba(2015)}]{diederik-kingma-2015-Adam}
Diederik~P Kingma and Jimmy Ba. 2015.
\newblock Adam: A method for stochastic optimization.
\newblock In \emph{Proceedings of the 3rd International Conference for Learning
  Representations}.

\bibitem[{K{\"u}{\c{c}}{\"u}k and Can(2020)}]{kuccuk2020survey}
Dilek K{\"u}{\c{c}}{\"u}k and Fazli Can. 2020.
\newblock Stance detection: A survey.
\newblock \emph{ACM Computing Surveys (CSUR)}, 53(1):1--37.

\bibitem[{Lempert(2008)}]{lempert2008poetics}
Michael Lempert. 2008.
\newblock The poetics of stance: Text-metricality, epistemicity, interaction.
\newblock \emph{Language in Society}, 37(4):569--592.

\bibitem[{Li and Caragea(2021)}]{li2021targetAug}
Yingjie Li and Cornelia Caragea. 2021.
\newblock Target-aware data augmentation for stance detection.
\newblock In \emph{Proceedings of the 2021 Conference of the North American
  Chapter of the Association for Computational Linguistics: Human Language
  Technologies}, pages 1850--1860.

\bibitem[{Li et~al.(2021)Li, Sosea, Sawant, Nair, Inkpen, and
  Caragea}]{li2021pstance}
Yingjie Li, Tiberiu Sosea, Aditya Sawant, Ajith~Jayaraman Nair, Diana Inkpen,
  and Cornelia Caragea. 2021.
\newblock P-stance: A large dataset for stance detection in political domain.
\newblock In \emph{Findings of the Association for Computational Linguistics:
  ACL-IJCNLP 2021}, pages 2355--2365.

\bibitem[{Liu et~al.(2019)Liu, Ott, Goyal, Du, Joshi, Chen, Levy, Lewis,
  Zettlemoyer, and Stoyanov}]{liu2019roberta}
Yinhan Liu, Myle Ott, Naman Goyal, Jingfei Du, Mandar Joshi, Danqi Chen, Omer
  Levy, Mike Lewis, Luke Zettlemoyer, and Veselin Stoyanov. 2019.
\newblock Roberta: A robustly optimized bert pretraining approach.
\newblock \emph{arXiv preprint arXiv:1907.11692}.

\bibitem[{Mohammad et~al.(2016)Mohammad, Kiritchenko, Sobhani, Zhu, and
  Cherry}]{mohammad-etal-2016-semeval}
Saif Mohammad, Svetlana Kiritchenko, Parinaz Sobhani, Xiaodan Zhu, and Colin
  Cherry. 2016.
\newblock \href {https://doi.org/10.18653/v1/S16-1003} {{S}em{E}val-2016 task
  6: Detecting stance in tweets}.
\newblock In \emph{Proceedings of the 10th International Workshop on Semantic
  Evaluation ({S}em{E}val-2016)}, pages 31--41, San Diego, California.
  Association for Computational Linguistics.

\bibitem[{Nguyen et~al.(2020)Nguyen, Vu, and Nguyen}]{nguyen2020bertweet}
Dat~Quoc Nguyen, Thanh Vu, and Anh-Tuan Nguyen. 2020.
\newblock Bertweet: A pre-trained language model for english tweets.
\newblock In \emph{Proceedings of the 2020 Conference on Empirical Methods in
  Natural Language Processing: System Demonstrations}, pages 9--14.

\bibitem[{Simaki et~al.(2018)Simaki, Simakis, Paradis, and
  Kerren}]{simaki2018detection}
Vasiliki Simaki, Panagiotis Simakis, Carita Paradis, and Andreas Kerren. 2018.
\newblock Detection of stance-related characteristics in social media text.
\newblock In \emph{Proceedings of the 10th Hellenic Conference on Artificial
  Intelligence}, pages 1--7.

\bibitem[{Uebersax(1982)}]{uebersax1982generalized}
John~S Uebersax. 1982.
\newblock A generalized kappa coefficient.
\newblock \emph{Educational and Psychological Measurement}, 42(1):181--183.

\bibitem[{Vaswani et~al.(2017)Vaswani, Shazeer, Parmar, Uszkoreit, Jones,
  Gomez, Łukasz Kaiser, and Polosukhin.}]{Ashish-2017-Transformer}
Ashish Vaswani, Noam Shazeer, Niki Parmar, Jakob Uszkoreit, Llion Jones,
  Aidan~N Gomez, Łukasz Kaiser, and Illia Polosukhin. 2017.
\newblock Attention is all you need.
\newblock In \emph{In Advances in Neural Information Processing Systems}, pages
  6000--6010.

\bibitem[{Vijayaraghavan et~al.(2016)Vijayaraghavan, Sysoev, Vosoughi, and
  Roy}]{vijayaraghavan2016deepstance}
Prashanth Vijayaraghavan, Ivan Sysoev, Soroush Vosoughi, and Deb Roy. 2016.
\newblock Deepstance at semeval-2016 task 6: Detecting stance in tweets using
  character and word-level cnns.
\newblock \emph{arXiv preprint arXiv:1606.05694}.

\bibitem[{Wang et~al.(2016)Wang, Huang, Zhu, and Zhao}]{wang2016ATAE}
Yequan Wang, Minlie Huang, Xiaoyan Zhu, and Li~Zhao. 2016.
\newblock Attention-based lstm for aspect-level sentiment classification.
\newblock In \emph{Proceedings of the 2016 conference on empirical methods in
  natural language processing}, pages 606--615.

\bibitem[{Xi et~al.(2020)Xi, Ma, Liou, Steinert-Threlkeld, Anastasopoulos, and
  Joo}]{xi2020understanding}
Nan Xi, Di~Ma, Marcus Liou, Zachary~C Steinert-Threlkeld, Jason Anastasopoulos,
  and Jungseock Joo. 2020.
\newblock Understanding the political ideology of legislators from social media
  images.
\newblock In \emph{Proceedings of the international aaai conference on web and
  social media}, volume~14, pages 726--737.

\bibitem[{Yuan et~al.(2022)Yuan, Zhao, Lu, and Qin}]{yuan2022ssr}
Jianhua Yuan, Yanyan Zhao, Yanyue Lu, and Bing Qin. 2022.
\newblock Ssr: Utilizing simplified stance reasoning process for robust stance
  detection.
\newblock In \emph{Proceedings of the 29th International Conference on
  Computational Linguistics}, pages 6846--6858.

\bibitem[{Zarrella and Marsh(2016)}]{zarrella2016mitre}
Guido Zarrella and Amy Marsh. 2016.
\newblock Mitre at semeval-2016 task 6: Transfer learning for stance detection.
\newblock \emph{arXiv preprint arXiv:1606.03784}.

\bibitem[{Zhou et~al.(2017)Zhou, Cristea, and Shi}]{zhou2017connecting}
Yiwei Zhou, Alexandra~I Cristea, and Lei Shi. 2017.
\newblock Connecting targets to tweets: Semantic attention-based model for
  target-specific stance detection.
\newblock In \emph{International Conference on Web Information Systems
  Engineering}, pages 18--32. Springer.

\bibitem[{Ziems et~al.(2023)Ziems, Held, Shaikh, Chen, Zhang, and
  Yang}]{ziems2023can}
Caleb Ziems, William Held, Omar Shaikh, Jiaao Chen, Zhehao Zhang, and Diyi
  Yang. 2023.
\newblock Can large language models transform computational social science?
\newblock \emph{arXiv preprint arXiv:2305.03514}.

\end{thebibliography}
\bibliographystyle{acl_natbib}

\newpage

\appendix

% \section{Appendix}
% \label{sec:appendix}

\begin{table*}[ht]
% \linespread{1.0}
    \centering
    \small
    \resizebox{0.96\linewidth}{!}{
    % \begin{tabular}{m{.3\textwidth}m{.7\textwidth}}
    \begin{tabular}{p{4.2cm}p{10cm}}
    \toprule
    \textbf{Environment Details} \\
    \midrule
        GPU Model & Single Tesla A100 with 40 GB memory; CUDA version 10.1. \\
        Library Version & Pytorch==1.8.1; Transformers==4.8.2. \\
        Computational Cost & Average 1.5 hours training time for one round. Average 3 rounds for each reported result (calculating mean of the result scores). \\
         \midrule
         \midrule
         \textbf{Stance Classification} &  \textbf{Experimental Configuration} \\
         \midrule
         Corpus & The datasets we used for training and evaluation are from published works \cite{mohammad-etal-2016-semeval,allaway2020VAST,li2021pstance,glandt-2021-tweetCovid} with the Creative Commons Attribution 4.0 International license. \\ 
         Pre-Processing & All samples are in English, and only for research use. Upper-case, special tokens, and hashtags are retained. \\
         \midrule
         RoBERTa-base & RoBERTa-base \cite{liu2019roberta} \\
         & Base Model: Transformer (12-layer, 768-hidden, 12-heads, 125M parameters). \\
         & Learning Rate: 2e-5, AdamW Optimizer, Linear Scheduler: 0.9. \\
         BERTweet & BERTweet-base \cite{nguyen2020bertweet} \\
         & Base Model: Transformer (12-layer, 768-hidden, 12-heads, 130M parameters). \\
         & Learning Rate: 2e-5, AdamW Optimizer, Linear Scheduler: 0.9. \\
         ATAE & ATAE-LSTM \cite{wang2016ATAE} \\
         & Base Model: Bi-LSTM (2-layer Bi-directional LSTM, hidden dimension is 300, linear calculation of target-level attention, 15M parameters). \\
         & Learning Rate: 3e-4, Adam Optimizer, Word Embedding: GloVe-840B. \\
         Product-of-Expert (PoE) & Product-of-Expert \cite{clark-etal-2019-PoE} \\
         & Base Model: RoBERTa (12-layer, 768-hidden, 12-heads, 125M parameters). \\
         & The bias-only model is trained on the tweet text without specified targets. \\
         & Learning Rate: 2e-5, AdamW Optimizer, Linear Scheduler: 0.9. \\
         \bottomrule
    \end{tabular}}
    \caption{Details of the experimental environment and the hyper-parameter setting.}
    \label{table-paramater-detail-appendix}
\vspace{-2.5cm}
\end{table*}

\begin{table*}[ht]
% \linespread{1.0}
    \centering
    \small
    \resizebox{0.96\linewidth}{!}{
    % \begin{tabular}{m{.3\textwidth}m{.7\textwidth}}
    \begin{tabular}{p{4.2cm}p{10cm}}
    \toprule
     \textbf{Annotation Enrichment} &  \textbf{} \\
     \midrule
     Original Dataset & The dataset we used for annotation enrichment is from published work \cite{mohammad-etal-2016-semeval} with the Creative Commons Attribution 4.0 International license. \\ 
     Pre-Processing & All samples are in English, and only for research use. Upper-case, special tokens, and hashtags are retained. The original sample size is 2,914. We filter out the samples where the specified target is explicitly mentioned in the text. Next, to obtain explicit objects in an extractive manner, we apply an off-the-shelf constituency parsing model, and collect all noun phrases in the constituency tree. \\
     \midrule
     Annotator Information & Four linguistic experts who are employed as full-time staff for natural language processing research participate the task. Their major language is English. The gender distribution covers female and male. \\
     Annotation Instruction & Each sample is a piece of text and a specified target, which forms one row in an excel file. Participants are asked to annotate the stance of the text author toward the specified target. The stance label is 3-class: \textit{Favor}, \textit{Against}, and \textit{None}. We introduce the stance detection task, and show some examples to all participants as preparation.
     In addition, there are two automatically-generated attributes: explicit mention and label alignment, which can be calculated after the manual stance labeling.\\
     Data Statistics & The enriched set contains 1,500 sample pairs, where each sample has annotation upon two different targets, and the adversarial pair size is 1.1k.\\
     Data Availability & Upon acceptance, following the previous published work \cite{mohammad-etal-2016-semeval}, the data can be accessed with the Creative Commons Attribution 4.0 International license, and only for research use.\\
     \bottomrule
    \end{tabular}}
    \caption{Details of the strategically-enriched data annotation.}
    \label{table-annotation-detail-appendix}
% \vspace{-1.5cm}
\end{table*}

\clearpage

\begin{table*}[ht]
\centering
\small
\resizebox{0.98\linewidth}{!}
{
\begin{tabular}{p{3.1cm}|p{1.4cm}<{\centering}p{1.4cm}<{\centering}p{1.4cm}<{\centering}|p{1.4cm}<{\centering}p{1.4cm}<{\centering}p{1.4cm}<{\centering}}
\toprule
 \textbf{Model}: RoBERTa-base &  \multicolumn{3}{c}{\textbf{In-Domain \& In-Target} (UB.)} & \multicolumn{3}{c}{\textbf{Single Corpus Training}} \\
\textbf{Test Set}  & \textbf{F1} & \textbf{Precision} & \textbf{Recall} & \textbf{F1} & \textbf{Precision} & \textbf{Recall} \\
\midrule
SemEval-16 Task-6 A  & 0.7023 & 0.7047 & 0.7318 & 0.7023 & 0.7047 & 0.7318 \\
SemEval-16 Task-6 B  & - & - & - & 0.3143 & 0.5126 & 0.2814 \\
P-Stance & 0.7745 & 0.7538 & 0.7977 & 0.4436 & 0.6597 & 0.5118 \\
VAST & 0.6661 & 0.6637 & 0.6850 & 0.3905 & 0.4192 & 0.3906 \\
Tweet-COVID  & 0.7244 & 0.7057 & 0.7500 & 0.2575 & 0.4377 & 0.1966 \\
\midrule
\midrule
 \textbf{Model}: RoBERTa-base &  \multicolumn{3}{c}{\textbf{Only Enriched Train Set}} & \multicolumn{3}{c}{\textbf{Adding Enriched Train Set}} \\
\textbf{Test Set}  & \textbf{F1} & \textbf{Precision} & \textbf{Recall} & \textbf{F1} & \textbf{Precision} & \textbf{Recall} \\
\midrule
SemEval-16 Task-6 A & 0.7088 & 0.7029 & 0.7360 & 0.7264 & 0.7286 & 0.7468 \\
SemEval-16 Task-6 B & 0.6099 & 0.6337 & 0.5883 & 0.6463 & 0.7342 & 0.5956 \\
P-Stance & 0.6764 & 0.6704 & 0.7054 & 0.6844 & 0.6689 & 0.7027 \\
VAST & 0.5798 & 0.6437 & 0.5496 & 0.5826 & 0.7167 & 0.4908 \\
Tweet-COVID & 0.4184 & 0.5043 & 0.4147 & 0.4579 & 0.5595 & 0.4146 \\
\bottomrule
\end{tabular}
}
\caption{\label{table-result-2-class}Results of the 2-class stance classification on multiple corpora. Macro-averaged F1, Precision, and Recall scores are reported. \textit{UB.} denotes the upper bound result from in-domain and in-target training on each corpus. Some examples of model prediction are shown in Appendix Table \ref{table-appendix-examples}.}
\vspace{-2.0cm}
\end{table*}

\begin{table*}[ht]
\centering
\small
\resizebox{0.98\linewidth}{!}
{
\begin{tabular}{p{3.0cm}|p{1.4cm}<{\centering}p{1.4cm}<{\centering}p{1.4cm}<{\centering}|p{1.4cm}<{\centering}p{1.4cm}<{\centering}p{1.4cm}<{\centering}}
\toprule
 &  \multicolumn{3}{c}{\textbf{Single Corpus Training}} & \multicolumn{3}{c}{\textbf{Adding Enriched Train Set}} \\
\textbf{Test Set}  & \textbf{ATAE} & \textbf{PoE} & \textbf{BERTweet} & \textbf{ATAE} & \textbf{PoE} & \textbf{BERTweet} \\
\midrule
SemEval16 Task-6 A & 0.6207 & 0.6531 & 0.7117 & 0.6266 & 0.6774 & 0.7347 \\
SemEval16 Task-6 B & 0.1743 & 0.4412 & 0.5208 & 0.2696 & 0.6051 & 0.6083 \\
P-Stance & 0.4129 & 0.5173 & 0.5742 & 0.4742 & 0.6831 & 0.6463 \\
VAST & 0.3134 & 0.4178 & 0.3604 & 0.3388 & 0.5532 & 0.5128 \\
Tweet-COVID & 0.2362 & 0.3124 & 0.4151 & 0.3784 & 0.3853 & 0.4951 \\
\bottomrule
\end{tabular}
}
\caption{\label{table-result-2-class-other-models}Various model performance of the 2-class stance classification on multiple corpora. Macro-averaged F1 scores are reported.}
\vspace{-2.0cm}
\end{table*}

\begin{table*}[ht]
% \linespread{1.0}
    \centering
    \small
    \resizebox{1.0\linewidth}{!}{
    \begin{tabular}{p{5.0cm}p{5.0cm}p{4.0cm}}
    \toprule
    \multicolumn{3}{l}{\textbf{Multi-Target Prediction Examples}} \\
    \midrule
    \multicolumn{3}{l}{\textbf{Text}:Considering the fact that Bush was a president of this country, I do not see it a joke that Trump is running ! \#Election2016} \\
    \textbf{Given Target}: ``Bush'' & \textbf{RoBERTa-base}: Against \scalebox{0.9}{\CheckmarkBold} & \textbf{Enhanced Model}: Against \scalebox{0.9}{\CheckmarkBold} \\
    \textbf{Given Target}: ``Donald Trump'' & \textbf{RoBERTa-base}: Against \scalebox{0.9}{\XSolidBrush} & \textbf{Enhanced Model}: Favor \scalebox{0.9}{\CheckmarkBold} \\
    \midrule
    \multicolumn{3}{l}{\textbf{Text}: 
    If @SpeakerPelosi wants to keep her job, she will fix this election as we all know its rigged. First, Tom Perez must go...} \\ 
    \multicolumn{3}{l}{and Bernie is our president. How can a news station air results when the primary that didnt even happen yet?} \\
    \textbf{Given Target}: ``Pelosi'' & \textbf{RoBERTa-base}: Against \scalebox{0.9}{\CheckmarkBold} & \textbf{Enhanced Model}: Against \scalebox{0.9}{\CheckmarkBold} \\
    \textbf{Given Target}: ``Bernie Sanders'' & \textbf{RoBERTa-base}: Against \scalebox{0.9}{\XSolidBrush} & \textbf{Enhanced Model}: Favor \scalebox{0.9}{\CheckmarkBold} \\
    \midrule
    \multicolumn{3}{l}{\textbf{Text}: 
    Why not? This protects both the officer and the civilian and it keeps things transparent. Then it would not be simply a } \\ 
    \multicolumn{3}{l}{matter of opinion when things go awry. It will be on videotape. BUT how much will it cost to store all this data and for how long?} \\
    \textbf{Given Target}: ``bodycamera'' & \textbf{RoBERTa-base}: Against \scalebox{0.9}{\XSolidBrush} & \textbf{Enhanced Model}: Favor \scalebox{0.9}{\CheckmarkBold} \\
    \textbf{Given Target}: ``videotape'' & \textbf{RoBERTa-base}: Against \scalebox{0.9}{\XSolidBrush} & \textbf{Enhanced Model}: Favor \scalebox{0.9}{\CheckmarkBold} \\
    \midrule
    \multicolumn{3}{l}{\textbf{Text}: 
    Why can’t people take this virus seriously and wear a damn mask? I can’t comprehend the childish behavior of some} \\ 
    \multicolumn{3}{l}{people who refuse to wear one. I just can’t.} \\
    \textbf{Given Target}: ``wear face masks'' & \textbf{RoBERTa-base}: None \scalebox{0.9}{\XSolidBrush} & \textbf{Enhanced Model}: Favor \scalebox{0.9}{\CheckmarkBold} \\
    \textbf{Given Target}: ``refuse to wear one'' & \textbf{RoBERTa-base}: None \scalebox{0.9}{\XSolidBrush} & \textbf{Enhanced Model}: Against \scalebox{0.9}{\CheckmarkBold} \\
    \bottomrule
    \end{tabular}}
    \caption{Examples of the prediction with and without enriched data annotation. Correct and incorrect predictions are indicated with the \scalebox{0.9}{\CheckmarkBold} and \scalebox{0.9}{\XSolidBrush} symbol, respectively.}
    \label{table-appendix-examples}
\vspace{-1.0cm}
\end{table*}

\end{document}